\documentclass[letterpaper]{article} 
\usepackage{aaai2026}
\usepackage{times}  
\usepackage{helvet}  
\usepackage{courier}  
\usepackage[hyphens]{url}  
\usepackage{graphicx} 
\usepackage{natbib}  
\usepackage{caption} 
\usepackage{algorithm}
\usepackage{algorithmic}

\usepackage{newfloat}
\usepackage{listings}
\DeclareCaptionStyle{ruled}{labelfont=normalfont,labelsep=colon,strut=off} 
\floatstyle{ruled}
\newfloat{listing}{tb}{lst}{}
\floatname{listing}{Listing}
\usepackage{amsmath}
\usepackage{cleveref}
\usepackage[T1]{fontenc}
\usepackage{booktabs}
\usepackage{multirow}
\usepackage{array}
\usepackage{pifont}
\usepackage{placeins}
\usepackage[most]{tcolorbox}
\usepackage{afterpage}

\newcounter{templatebox}

\newtcolorbox{mytbox}[2][]{%
  enhanced,
  float=t,
  colback=gray!10,
  colframe=gray!50,
  fonttitle=\bfseries,
  sharp corners=southwest,
  before title={\refstepcounter{templatebox}},  
  title={Template~\thetemplatebox: #2},         
  #1                                            
}

\title{Measuring What LLMs Think They Do: SHAP Faithfulness and Deployability on Financial Tabular Classification}

\author{
    Saeed AlMarri\textsuperscript{\rm 1, \rm 2},
    Mathieu Ravaut\textsuperscript{\rm 2},
    Kristof Juhasz\textsuperscript{\rm 2},
    Gautier Marti\textsuperscript{\rm 2},
    Hamdan Al Ahbabi\textsuperscript{\rm 1, \rm 2},
    Ibrahim Elfadel\textsuperscript{\rm 1}
}

\affiliations{
    \textsuperscript{\rm 1}Khalifa University, Abu Dhabi, United Arab Emirates\\
    \textsuperscript{\rm 2}Abu Dhabi Investment Authority (ADIA), Abu Dhabi, United Arab Emirates\\
}

\begin{document}

\maketitle

\begin{abstract}
   Large Language Models (LLMs) have attracted significant attention for classification tasks, offering a flexible alternative to trusted classical machine learning models like LightGBM through zero-shot prompting.
However, their reliability for structured tabular data remains unclear, particularly in high-stakes applications like financial risk assessment.
Our study systematically evaluates LLMs and generates their SHAP values on financial classification tasks. 
Our analysis shows a divergence between LLMs self-explanation of feature impact and their SHAP values, as well as notable differences between LLMs and LightGBM SHAP values.
These findings highlight the limitations of LLMs as standalone classifiers for structured financial modeling, but also instill optimism that improved explainability mechanisms coupled with few-shot prompting will make LLMs usable in risk-sensitive domains. \\
\end{abstract}


\section{Introduction}
\label{sec:intro}

Chatbots powered by Large Language Models (LLMs) such as GPT-4 \citep{achiam2023gpt}, have demonstrated strong performance across a range of natural language processing (NLP) tasks, including classification and reasoning \citep{wei2022chain}. Their ability to function as classifiers without explicit training pipelines, relying solely on few-shot or zero-shot prompting, has gained significant attention \citep{brown2020language,qin2023chatgpt}. This raises fundamental questions about the reliability and validity of LLM-based classification, particularly in comparison to classical machine learning models such as XGBoost \citep{chen2016xgboost} and LightGBM \citep{ke2017lightgbm}.

Traditional classification tasks require structured pipelines involving feature engineering, model training, validation, and hyperparameter tuning. Fine-tuning models on tabular data, in particular, demands finesse and expertise in data preprocessing, GPU management, and balancing class distributions to prevent trivial solutions. In contrast, LLMs bypass fine-tuning entirely, requiring only natural language prompting. This reduces technical barriers, making them accessible to non-experts - a valuable boost to adoption of these tools. Nonetheless, before entrusting LLMs with critical decisions, a question remains: \textbf{\emph{How to explain predictions from LLM classifiers?}}

This question is particularly relevant in finance, a high-stakes domain where transparency and accountability are critical because algorithm outputs directly affect credit access, interest rates and regulatory compliance \citep{doshi-velez2017towards}. Financial institutions operate under strict governance frameworks such as Basel III \citep{BaselIII} and GDPR \citep{gdpr-2016}, where opaque risk assessment models can lead to regulatory breaches, reputational damage, and unfair or discriminatory decisions, causing trust concerns. Unlike decision trees or gradient boosting models, LLMs are complex black-box models with billions of parameters, making interpretability a key challenge. This has led to increasing interest in Explainable AI (XAI) techniques to analyze LLMs' internal logic and assess their alignment with human-interpretable decision patterns.

In this study, we investigate LLMs' capacity to introspect and explain their own predictive mechanism. For LLM explainability, we employ Shapley Additive Explanations (SHAP) \citep{lundberg2017}, for which we provide an efficient LLM implementation. To explain the prediction mechanism, we prompt LLMs to self-explain on the impact of each feature on the classification task. From a deployability perspective, we go beyond accuracy to assess the faithfulness of explanations, their sensitivity to prompt and serialization variations, and the feasibility of post-hoc auditing under regulatory expectations for high-risk financial AI systems.

Across four open-source LLMs and three binary classification tasks with financial tabular data, our experiments show that overall, zero-shot LLMs are poorly aware of their predictive mechanism, as their self-explanations do not align with their SHAP values. LLMs SHAP values also highly differ from those of LightGBM. Additional work is needed to improve usability of these LLMs in the financial domain, such as model augmentation \citep{theuma2024equipping}, or more elaborate inference pipelines involving few-shot prompting.

\section{Related Work}
\label{sec:related}


% ...



\begin{mytbox}[label=box:template1]{Instance-Level Prompt}
\small
Predict \texttt{<Task Description>}. Use the features provided below to assess the likelihood of \texttt{<Positive Class Name>}.

\vspace{2mm}
\textbf{\texttt{<Task Name>} Details:}
\begin{verbatim}
<feature_1 name>: <feature_1 value>
...
<feature_N name>: <feature_N value>
\end{verbatim}

\vspace{2mm}
Provide your estimated probability of the \texttt{<Positive Class Name>}.
Do NOT perform coding or calculations, just provide the probability.

Your answer should only contain the probability estimate in JSON:
\begin{verbatim}
{
  "Estimated <Positive Class Name>": 
  <float value between 0 and 1>
}
\end{verbatim}
\end{mytbox}

\subsection{LLMs for Tabular Data}

The application of LLMs to tabular data has emerged as a novel approach in regression tasks. Unlike traditional machine learning (ML) models, which require explicit training on labeled datasets, LLMs can be prompted with feature sets in a zero-shot manner, eliminating task-specific training. This method involves serializing tabular data into a natural language format and leveraging the LLM's pre-trained knowledge to make predictions.

\citet{hegselmann2023} introduce TabLLM, a framework that utilizes LLMs for few-shot classification of tabular data by converting rows into natural language representations and providing a brief description of the classification problem. Their findings suggest that LLMs can outperform traditional deep learning models in certain tabular classification tasks \cite{hegselmann2023}.
Similarly, \citet{shi2024} propose Zero-shot Encoding for Tabular data with LLMs (ZET-LLM), an approach that treats auto-regressive LLMs as feature embedding models for tabular prediction tasks. By implementing a feature-wise serialization and addressing challenges like limited token lengths and missing data, they demonstrated that LLMs could serve as effective zero-shot feature extractors without fine-tuning \cite{shi2024}.

In this study, we leverage LLMs as zero-shot classifiers on three financial datasets, directly injecting feature names and feature values in the prompt.

\subsection{LLMs Explainability}

Feature attribution methods such as \textit{SHAP} \citep{lundberg2017} are widely used to assess feature importance in classical machine learning models like XGBoost \citep{chen2016xgboost}, LightGBM \citep{ke2017lightgbm} or CatBoost. However, their role in LLM-based classification remains underexplored, largely due to the high computational cost: SHAP requires a high number of inference passes. 

TokenSHAP \citep{goldshmidt2024tokenshap} combines cooperative game theory framework with efficient token attribution. Through Monte Carlo sampling, it estimates each token's SHAP contribution to the prediction. 
\citet{mohammadi2024explaining} reduced the input space by using a fixed prompt template dissected into segments.

Our study is the first to compute SHAP-based feature importance on LLMs prompted to predict a probabilistic outcome on structured financial classification datasets.

\begin{mytbox}[label=box:template2]{Feature-Level Prompt}
\small
You are working on predicting <Task Description>.
It is a binary classification task where the positive class corresponds to <Positive Class Name>.

One of the features is the following: \begin{verbatim} <feature name> \end{verbatim}

What impact do you think this feature will have on the classification task?
Provide your answer among 3 possible strings: positive | neutral | negative. 
Do NOT output reasoning or explanations, just output the feature impact string.
Write it in the following format in JSON:

\begin{verbatim}
{
  "Feature impact": <string among 
     positive | negative | neutral>,
}
\end{verbatim}
\end{mytbox}

\begin{mytbox}[label=box:template3]{Feature-Level Prompt with Self-Explanation}
\small
You are working on predicting <Task Description>.
It is a binary classification task where the positive class corresponds to <Positive Class Name>.

One of the features is the following: \begin{verbatim} <feature name> \end{verbatim}

What impact do you think this feature will have on the classification task?
Provide your answer among 3 possible strings: positive | neutral | negative. 
Also provide a brief explanation of this feature's impact. 
Just output the feature impact string and your explanation.
Write it in the following format in JSON:

\begin{verbatim}
{
  "Feature impact": <string among 
     positive | negative | neutral>,
  "Explanation": <string value>
}
\end{verbatim}
\end{mytbox}

\subsection{LLMs Self-Explanations}

\begin{table*}[t]
\centering
\small
\resizebox{0.95\textwidth}{!}{%
\begin{tabular}{lcccccccc}
\toprule
& \multicolumn{2}{c}{\textbf{Bankruptcy}} 
& \multicolumn{2}{c}{\textbf{Loan Repayment}} 
& \multicolumn{2}{c}{\textbf{License Expiration}}
& \multicolumn{2}{c}{\textbf{Average}} \\
\cmidrule(lr){2-3}\cmidrule(lr){4-5}\cmidrule(lr){6-7}\cmidrule(lr){8-9}
\textbf{Model}
& ROC-AUC & PR-AUC
& ROC-AUC & PR-AUC
& ROC-AUC & PR-AUC
& ROC-AUC & PR-AUC \\
\midrule
\textbf{Gemma-2-9B}                  & \textbf{0.641} & 0.059 & \textbf{0.669} & \textbf{0.878} & \textbf{0.601} & \textbf{0.029} & \textbf{0.637} & \textbf{0.322} \\
\textbf{Llama-3.2-3B}            & 0.524 & 0.041 & 0.616 & 0.851 & 0.439 & 0.019 & 0.526 & 0.304 \\
\textbf{Qwen-2.5-7B}             & 0.630 & 0.054 & 0.591 & 0.839 & 0.433 & 0.018 & 0.551 & 0.304 \\
\textbf{Mistral-7B-v0.3}         & 0.624 & \textbf{0.060} & 0.651 & 0.873 & 0.573 & 0.026 & 0.614 & 0.320 \\
\bottomrule
\end{tabular}
}
\caption{LLMs classification performance summary. Bold numbers highlight best performance on each dataset.}
\label{tab:perf}
\end{table*}

\begin{table*}[t]
\centering
\small
\resizebox{0.95\textwidth}{!}{%
\begin{tabular}{lccccc}
\toprule
\textbf{LLM} 
& \textbf{Using rationale?}
& \textbf{Bankruptcy}
& \textbf{Loan Repayment}
& \textbf{License Expiration}
& \textbf{Average} \\
\midrule

\multirow{2}{*}{\textbf{Gemma-2-9B}}            
& \ding{55}  
& 50.0\% (10/20) 
& \textbf{66.7}\% (8/12) 
& 33.3\% (5/15) 
& 50.0\% \\
& \ding{51} 
& \textbf{65.0}\% (13/20) 
& \textbf{66.7}\% (8/12) 
& 40.0\% (6/15) 
& \textbf{57.2}\% \\

\midrule

\multirow{2}{*}{\textbf{Llama-3.2-3B}}    
& \ding{55} 
& 35.0\% (7/20) 
& 41.7\% (5/12) 
& \textbf{60.0}\% (9/15) 
& 45.6\% \\
& \ding{51} 
& 30.0\% (6/20) 
& 58.3\% (7/12) 
& 53.3\% (8/15) 
& 47.2\% \\

\midrule

\multirow{2}{*}{\textbf{Qwen-2.5-7B}}     
& \ding{55} 
& 15.0\% (3/20) 
& 33.3\% (4/12) 
& 26.7\% (4/15) 
& 25.0\% \\
& \ding{51} 
& 30.0\% (6/20) 
& 25.0\% (3/12) 
& 33.3\% (5/15) 
& 29.4\% \\
     
\midrule

\multirow{2}{*}{\textbf{Mistral-7B-v0.3}} 
& \ding{55} 
& 35.0\% (7/20) 
& 25.0\% (3/12) 
& 26.7\% (4/15) 
& 28.9\% \\
& \ding{51} 
& 20.0\% (4/20) 
& 58.3\% (7/12) 
& 33.3\% (5/15) 
& 37.2\% \\
 
\bottomrule
\end{tabular}
}
\caption{Percent features in agreement between LLMs self-explanation and LLMs SHAP values. Due to the three-class setup, the baseline accuracy is 33\%. Bold numbers highlight best performance on each dataset. The rationale column shows whether a prompt asking for self-explanation was presented. In parenthesis are shown the fraction of features which each LLM correctly predicted.}
\label{tab:self_explain}
\end{table*}

Another line of research focuses on LLM-generated \emph{rationales} or \emph{self-explanations} of their predictions. 

\citet{huang2023can} found that LLM rationales can be plausible but do not reflect internal reasoning. 
\citet{dehghanighobadi2025can} analyzed counterfactual explanations, showing that LLMs struggle with causal dependencies. 
\citet{sarkar2024large} argues that LLMs lack self-explanatory capabilities due to opaque training dynamics, while \citet{turpin2023language} showed that CoT-generated explanations \citep{wei2022chain} can be misleading.

A key question arising from this endeavour, and which we explore through this paper, is whether LLMs' self-explanations \emph{align} with actual feature contribution.

\section{Methodology}
\label{sec:methodology}

\subsection{Datasets}

We experiment with three classification tasks (each framed as a binary classification) covering vastly different aspects of financial machine learning.

\subsubsection{Bankruptcy}
\label{sec:3_1_1}

We use the Polish Companies Bankruptcy dataset, explored in \citet{zikeba2016ensemble}, keeping the subset with 1-year future bankruptcy prediction, totaling 7,027 companies including 271 going bankrupt (3.9\% positive ratio). To keep the features size manageable, we only keep the top 20 features out of the dataset's initial 64, identified by computing the feature importance of a LightGBM model. All 20 features are numeric, see \Cref{tab:bankruptcy_features} in \Cref{sec:appendix_features}.

\subsubsection{Loan Repayment}
\label{sec:3_1_2}

We use a very popular Kaggle dataset\footnote{\url{https://www.kaggle.com/datasets/sndpred/loan-data}}, with 79,206 loan applications, including 63,629 that are fully paid (80.3\% positive ratio). Of the 21 features, 12 are numeric, see \Cref{tab:loan_features} in \Cref{sec:appendix_features}.

\subsubsection{License Expiration}
\label{sec:3_1_3}

We leverage the recently introduced Hong Kong Securities and Futures Commission (SFC) dataset \citep{alketbi2024mapping}. Due to the size, we only keep the last month (January 2024), where 23,001 employees are recorded and 478 see their SFC license not renewed within the next month (2.1\% positive ratio). All 15 features used are numeric, see \Cref{tab:license_features} in \Cref{sec:appendix_features}.

\subsection{Models \& Inference}

We use four recent open-source LLMs: \texttt{gemma-2-9b-instruct} \citep[\textbf{Gemma-2-9B},][]{team2024gemma}, \texttt{llama-3.2-3b-instruct} \citep[\textbf{Llama-3.2-3B},][]{dubey2024llama}, \texttt{qwen-2.5-7b-instruct} \citep[\textbf{Qwen-2.5-7B},][]{yang2024qwen2} and \texttt{mistral-7b-instruct-v0.3} \citep[\textbf{Mistral-7B-v0.3},][]{jiang2023mistral7b}. Weights were downloaded from HuggingFace \citep{wolf2020transformers}, and inference was done locally through vLLM\footnote{\url{https://github.com/vllm-project/vllm}} on two NVIDIA A10G 24GB GPUs. The same instance‑level prompt was used for each type of LLM, shown in Template 1. The probability of the positive class was generated in JSON format.

\begin{figure*}[t!]
\centering
\includegraphics[width=0.98\textwidth]{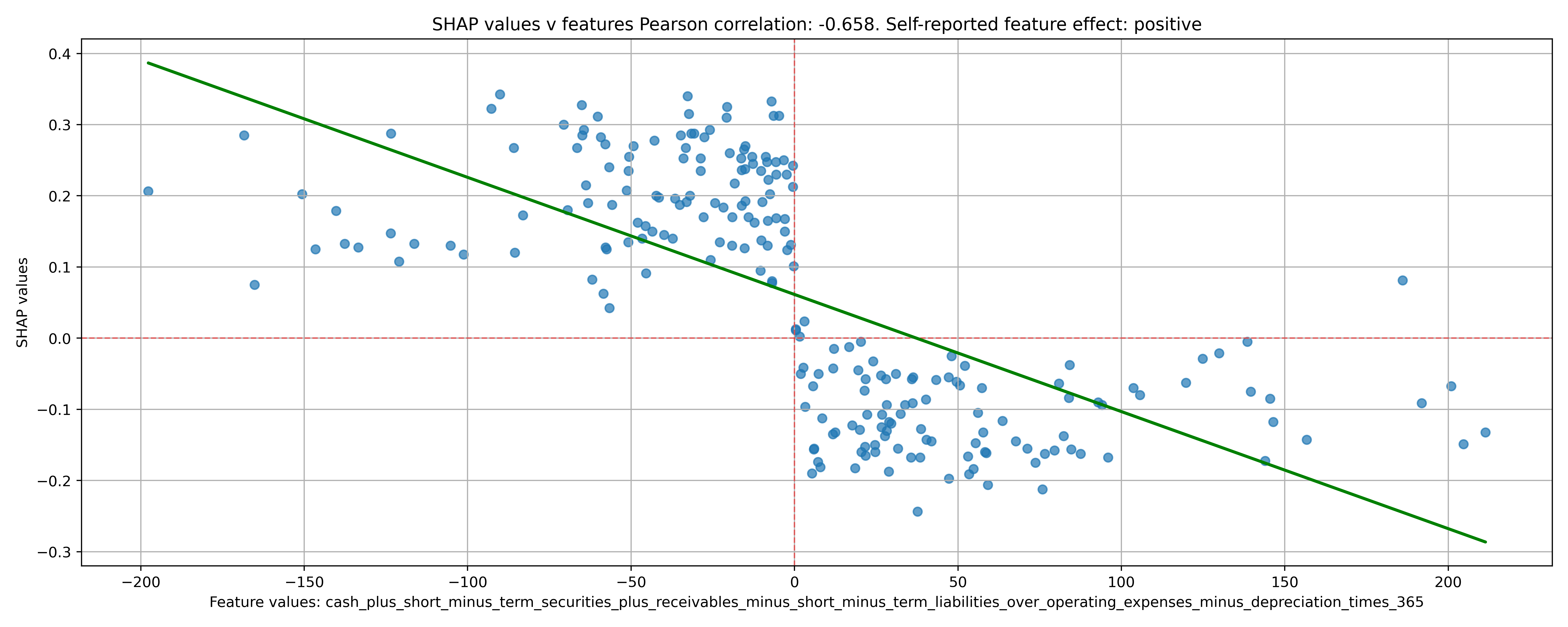}
\caption{SHAP dependence plot for Qwen-2.5-7B highest importance feature on the Bankruptcy dataset.}
\label{fig:qwen_highest}
\end{figure*}

\begin{figure*}[t!]
\centering
\includegraphics[width=0.98\textwidth]{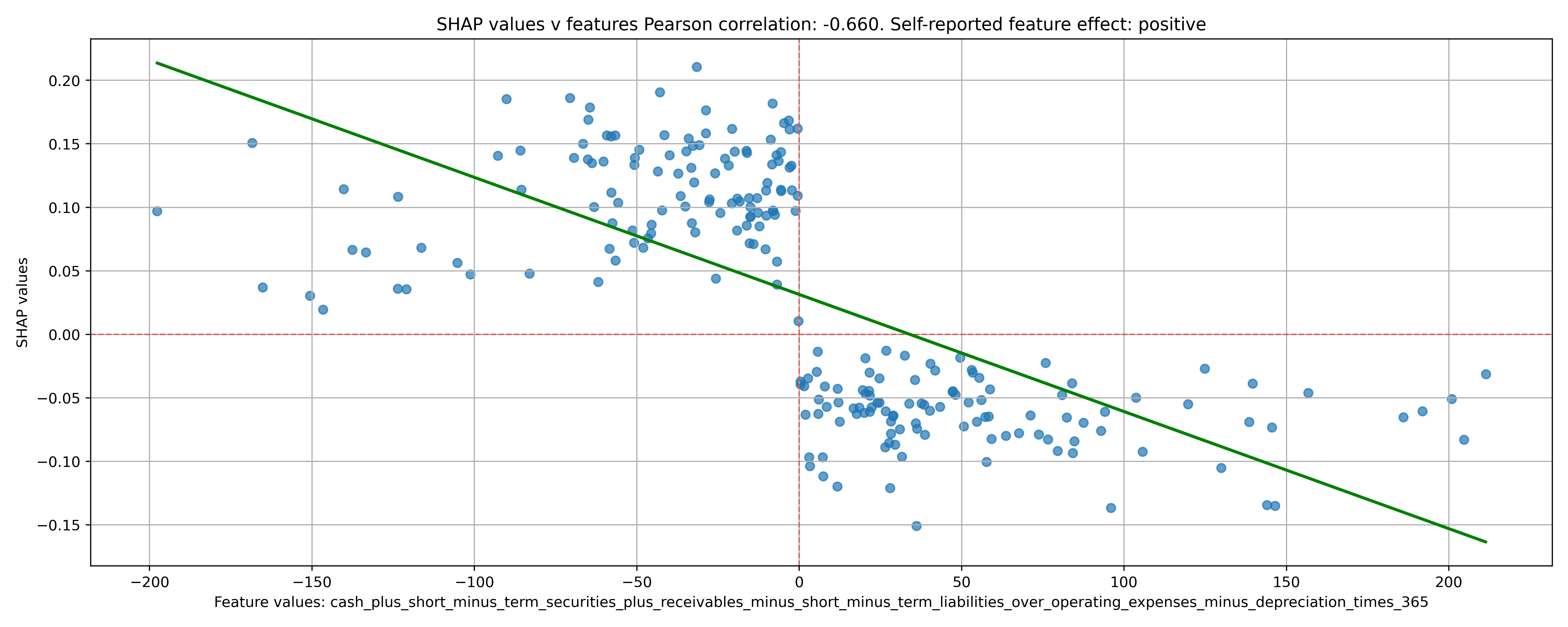}
\caption{SHAP dependence plot for Mistral-7B-v0.3 highest importance feature on the Bankruptcy dataset.}
\label{fig:mistral_highest}
\end{figure*}

\subsection{Explainability}

\subsubsection{LLMs SHAP Values}
\label{sec:3_3_1}

\begin{figure*}[htb]
   \centering
   \includegraphics[width=\textwidth]{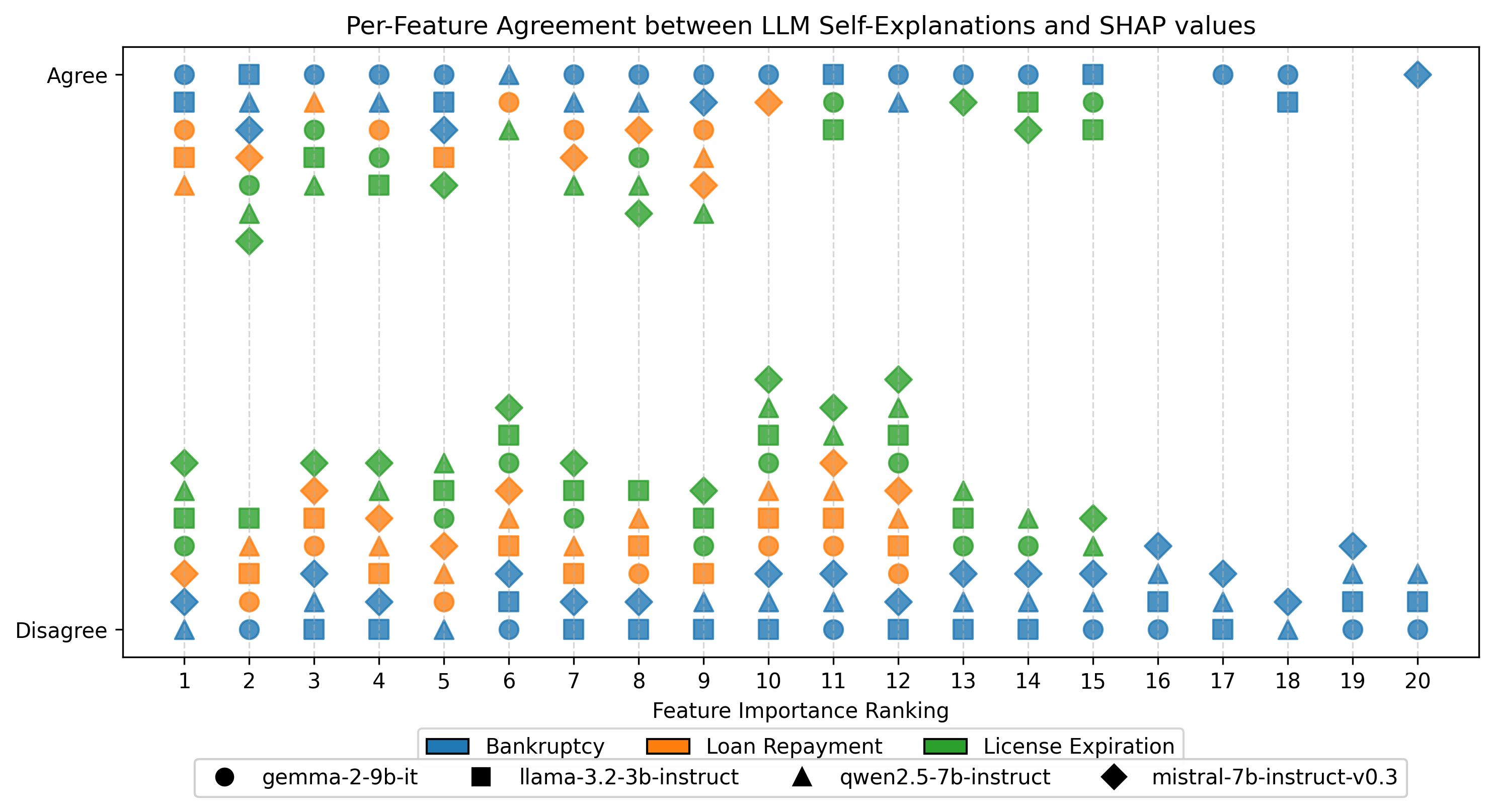}
   \caption{Per-feature agreement between LLMs self-explanations with rationale and LLMs SHAP values. Clear disagreements persist even among the top‑k important features}
   \label{fig:per_feature}
\end{figure*}

We use SHAP for post hoc explanations \citep{lundberg2017}, specifically the model‑agnostic \texttt{PermutationExplainer}.
We adopt this efficient SHAP estimator because our prediction function is an LLM inference, which is costly. 
To balance accuracy and runtime, we sample 250 instances from each dataset for explanation. 
We construct the background (masker) via \(k\)-means clustering with \(C=5\) centroids using \texttt{shap.kmeans}, and set the \texttt{max\_evals} budget, so the explainer executes exactly \(T=4\) permutations in our experiments.

\paragraph{Approximate cost (model calls).}
Let \(K\) be the number of instances explained, \(M\) the number of features (e.g. 20 on Bankruptcy), \(B\) the number of background draws per masked evaluation (here \(B=C=5\)), and \(T\) the number of random permutations. The \textbf{PermutationExplainer} requires approximately:

\begin{equation}
\#\text{calls} \;\approx\; K \times T \times (M+1) \times B \;=\; \mathcal{O}(K\,T\,M\,B)
\end{equation}

model evaluations. In SHAP’s implementation, \(T\) is governed by \texttt{max\_evals} via the practical rule:

\begin{equation}
T \;\approx\; \Big\lfloor \frac{\texttt{max\_evals}}{2M} \Big\rfloor
\end{equation}

i.e., roughly \(2M\) masked evaluations per permutation path. With \(\texttt{max\_evals}=200\) and \(M=21\), this yields \(T=\lfloor 200/(2\times 21)\rfloor=4\). Using \(B=5\), the per instance cost is therefore \(\approx 4 \times (21{+}1) \times 5 = 440\) in the model calls (for the loan repayment dataset with M = 21; other datasets scale accordingly).

\paragraph{Why it is more efficient than \texttt{KernelExplainer}?}
With a summarized background of \(C\) centroids, the dominant model‑call complexity of \texttt{KernelExplainer} scales as:

\begin{equation}
\#\text{calls} \;\approx\; K \times C \times M^2 \;=\; \mathcal{O}(K\,C\,M^2)
\end{equation}

due to sampling coalitions and fitting a kernel‑weighted regression. In our setting (\(C=5\), \(M=21\)) this is \(\approx 5 \times 21^2 = 2205\) evaluations per instance. By contrast, \texttt{PermutationExplainer} scales linearly in \(M\) and avoids the regression solve, yielding an expected per‑instance reduction of

\begin{equation}
\text{speedup} \;\approx\; \frac{C\,M^2}{T\,(M+1)\,B} \;\approx\; \frac{2205}{440} \;\approx\; 5\times
\end{equation}

The fivefold decrease in model calls translates into substantially lower LLM inference time while maintaining faithful attributions, which is why we use \texttt{PermutationExplainer} with \(T=4\).

\subsubsection{LLMs Self-Explanations}

Motivated by the new reasoning capabilities of LLMs \citep{wei2022chain,huang2022towards,ke2025survey}, we leverage the LLM as an explainability tool on its own \citep{huang2023can}. Specifically, for each feature, we prompt its description to the LLM and ask it to predict whether the feature will have a negative, neutral or positive impact on the classification ; with an option to provide a self-explanation (rationale) about its prediction. Template 2 and Template 3 show the two corresponding feature-level prompt templates.

\section{Experiments}
\label{sec:experiments}

Overall, the classification performance results are consistently above random chance. Because the models are used zero‑shot, performance is modest (\Cref{tab:perf}); nevertheless, results indicate \textit{detectable} signal on finance datasets. To us, this means that with some modification, LLMs have potential for being used in the financial domain. In the following analysis, we only consider numerical features when computing SHAP values.

\subsection{LLMs SHAP Values and LLMs Self-Explanations Comparison}

\Cref{fig:qwen_highest} and \Cref{fig:mistral_highest} demonstrate the SHAP dependence plots on the examples of the Qwen-2.5-7B and Mistral-7B-v0.3 models, showcasing the disparity between what LLMs think they do (positive or negative) vs. what they actually do (SHAP). In both cases, the LLM incorrectly predicts a positive feature impact whereas the SHAP values are strongly negatively correlated with the feature values.

To quantify this disparity, we compare SHAP values with the feature values through Pearson correlation coefficient. We classify the correlation into three feature impacts: negative (Pearson below -0.1), neutral (between -0.1 and 0.1) and positive (greater than 0.1). This feature impact is compared against the LLM self-explanation as a way to assess the LLM's own understanding of its classification process. 

Results are shown in \Cref{tab:self_explain}. 
The prompt asking for a rationale provides a moderate, yet consistent improvement in classification agreement with SHAP feature impact over the baseline prompt. Gemma-2-9B outperforms other LLMs both in terms of performance and self-explanation accuracy (\Cref{tab:perf,tab:self_explain}). However, even Gemma-2-9B scores only a bit above 50\% self-explanation accuracy on average across all datasets. 

\Cref{fig:per_feature} extends the analysis by showing the LLM self-explanation agreement for each individual, and sorting features by decreasing feature importance. At each feature importance rank, we split the agreement on each (dataset, model) pair in two buckets: Agree on top and Disagree on bottom. As seen, even for the top three most important features, which should be trivial to classify, there are many cases where LLMs cannot predict the correct feature impact on classification. 

Thus, we conclude that \textbf{zero-shot LLMs are not able to identify a feature's impact on classification}. To take advantage of the potential demonstrated in \Cref{tab:perf}, few-shot performance will need to be assessed.

\begin{table}[t]
\centering
\setlength{\tabcolsep}{1mm}
\resizebox{\columnwidth}{!}{%
\begin{tabular}{lcc cc cc}
\toprule
\multirow{2}{*}{\textbf{LLM}}
& \multicolumn{2}{c}{\textbf{Bankruptcy}}
& \multicolumn{2}{c}{\textbf{Loan}}
& \multicolumn{2}{c}{\textbf{License}} \\
\cmidrule(lr){2-3}\cmidrule(lr){4-5}\cmidrule(l){6-7}
 & $\tau$ 
 & Dir\% 
 & $\tau$ 
 & Dir\% 
 & $\tau$ 
 & Dir\% \\
\midrule

\textbf{Gemma-2-9B}            
& 0.011	
& 65.0\%
& 0.000	
& 50.0\%
& -0.352
& 66.7\% \\

\textbf{Llama-3.2-3B}
& 0.084	
& 50.0\%
& 0.276	
& 58.3\% 
& -0.276	
& 53.3\% \\

\textbf{Qwen-2.5-7B}
& -0.042	
& 50.0\% 
& 0.190	
& 75.0\% 
& 0.029	
& 60.0\% \\

\textbf{Mistral-7B-v0.3}
& 0.116	
& 55.0\% 
& 0.124	
& 58.3\% 
& -0.143	
& 53.3\% \\

\midrule

\textbf{Average}
& 0.042	
& 55.0\% 
& 0.148	
& 60.4\% 
& -0.186	
& 58.3\% \\

\bottomrule
\end{tabular}
}
\caption{
    Alignment between LLMs and LightGBM SHAP values. \textbf{$\tau$} is the Kendall rank correlation on full feature order, and \textbf{Dir\%} = \% of features with identical SHAP sign.
}
\label{tab:lightgbm}
\end{table}

\subsection{LLMs and LightGBM SHAP Values Comparison}

To further investigate LLMs SHAP Values, we compare them against the ones of LightGBM \citep{ke2017lightgbm}, a well-established, state-of-the-art gradient boosting decision tree model. Results displayed in \Cref{tab:lightgbm} show that LLMs and LightGBM have low correlation in terms of SHAP values. Their agreement on the direction of a feature's impact is just a bit above random chance on average (50-60\%). We conclude that LLMs' classification reasoning greatly differs from the classification process of LightGBM.



\subsection{Why do LLMs mis-sign top features?}
\label{sec:mis-sign}

\Cref{fig:qwen_highest} and \Cref{fig:mistral_highest} illustrate cases where LLM self-reported impacts (``positive'') contradict SHAP dependence trends (strongly \emph{negative}). \Cref{tab:self_explain} shows that asking for rationales increases agreement only modestly (e.g., Gemma-2-9B improves from 50.0\% to 57.2\% on average), and \Cref{tab:lightgbm} reports low Kendall's~$\tau$ alignment with LightGBM. Together, these suggest that LLM self-explanations are shaped by \emph{lexical priors} in feature names rather than the dataset-specific conditional relationships captured by SHAP.

\paragraph{Potential mitigation techniques (not requiring re-training):}
\begin{itemize}
  \item \textbf{Feature-name neutralization:} anonymize feature names (e.g., $f_1,\ldots,f_M$) in prompts to reduce bias from tokens such as \emph{cash}, \emph{profit}, or \emph{liabilities}, then map back for human consumption.
  \item \textbf{Serialization robustness:} vary feature order, delimiters, and descriptions to assess prediction/explanation stability; LLMs on tables are known to be sensitive to serialization choices \citep{hegselmann2023}.
  \item \textbf{Agreement reporting:} in addition to percent agreement, report beyond-chance measures (e.g., Cohen's $\kappa$ or Matthews correlation) between SHAP signs and LLM labels.
  \item \textbf{Sanity checks for explanations:} apply label or feature randomization tests to ensure explanations collapse under appropriate perturbations \citep{Adebayo2018}.
\end{itemize}

We view the persistent disagreements at top importance ranks (\Cref{fig:per_feature}) as a deployability red flag whenever feature-level justifications are required by policy or governance.

\subsection{Interpreting Performance Under Class Imbalance}

\paragraph{Why PR‑AUC matters here.} 
All three tasks are substantially imbalanced: bankruptcy (3.9\% positives), license expiration (2.1\%), and loan repayment (80.3\% positive class defined as fully paid). In such settings, PR‑AUC is more informative than ROC‑AUC because it directly reflects precision at given recalls and is sensitive to the positive rate, unlike ROC‑AUC which can appear optimistic when negatives dominate \citep{saito2015pr, davis2006}. 

\paragraph{Lift over baseline.}
To contextualize \Cref{tab:perf}, we report the \emph{PR-AUC lift} defined as $\text{Lift}=\text{PR-AUC}/\text{baseline}$. On Bankruptcy, PR-AUCs of respectively 0.059, 0.041, 0.054, 0.060 correspond to lifts of $\mathbf{1.51\times}$, $1.05\times$, $1.39\times$, and $\mathbf{1.54\times}$, respectively (average $1.37\times$) over the 3.9\% baseline. On Loan Repayment, PR-AUCs of respectively 0.878, 0.851, 0.839, 0.873 imply lifts of $1.09\times$, $1.06\times$, $1.05\times$, and $1.09\times$ (average $1.07\times$) against the 80.3\% baseline. On License Expiration, PR-AUCs of respectively 0.029, 0.019, 0.018, 0.026 yield lifts of $1.38\times$, $0.91\times$, $0.86\times$, and $1.24\times$ (average $1.10\times$) over the 2.1\% baseline. These numbers (from \Cref{tab:perf}) show weak but non-trivial signal on Bankruptcy and mixed results on the very sparse License task; Loan Repayment gains are small because the baseline (0.803) is already high.

\paragraph{Deployability implication.}
Even without any fine-tuning or feature engineering, zero-shot LLMs recover modest signal on some tabular finance tasks. However, the small deltas over baseline and the sub-baseline result in one License setting indicate that \emph{few-shot prompting}, \emph{ensembling}, or \emph{hybridization} with tabular models are likely prerequisites for deployment in risk-sensitive contexts \citep{hegselmann2023, shi2024}.

\section{Discussion}
\label{sec:discussion}

\subsection{Limitations and Threats to Validity}

Several limitations of this study warrant discussion. First, the interpretation of performance metrics under data imbalance presents challenges. We emphasized the use of PR-AUC and baseline-normalized lift, as relying solely on ROC-AUC can overstate apparent gains on highly skewed datasets \cite{saito2015pr}. Second, our SHAP-based attribution analysis depends on the choice of background samples and estimator. For computational efficiency, we employed the \textit{PermutationExplainer} with k-means maskers, but attributions may vary with both the background distribution and the \texttt{max\_evals} parameter. A small ablation varying these choices would further strengthen interpretability claims.

Third, LLM outputs exhibit sensitivity to prompt serialization, that is, variations in feature order or phrasing can affect the model’s reasoning trace. Future work should systematically quantify this prompt sensitivity through controlled perturbation benchmarks such as SUC or serialization robustness tests. Fourth, the computational cost of explainability presents a practical constraint. We explained $K=250$ rows per dataset using SHAP, with $T=4$ and $B=5$, leading to approximately 110k model calls per dataset-model pair. Across three datasets and four LLMs, this results in roughly 1.32 million evaluations, which poses scalability challenges for deployment-grade auditing.

Finally, our analysis highlights that LLM self-explanations should not be equated with true causal mechanisms. The low agreement observed between SHAP attributions and LLM-generated rationales (see Tables 2–3) indicates that while LLM rationales may appear plausible, they often lack faithfulness to underlying model behavior. Consequently, explanation outputs must be independently audited, for instance through sanity checks or falsification tests, before being considered reliable in regulated decision-making contexts.

\subsection{Calibration and Decision-Theoretic Concerns.} 

Since LLMs output \textit{probabilities}, deployment must prioritize calibration. Classic calibration methods, such as Platt scaling or isotonic regression, trained on validation sets can enhance downstream decision quality \cite{niculescuMizil2005, zadrozny2002}. Future work should consider (i) reporting reliability diagrams and Brier scores; (ii) providing 95\% CIs for AUC and PR-AUC; (iii) translating metrics into cost-sensitive operating points. These calibration-aware strategies are critical, especially
in asymmetric cost domains such as financial services

\subsection{Deployment Implications}

\subsubsection{When (Not) to Use Zero-Shot LLMs for Tabular Finance.} 

Our findings align with standard deployability dimensions: \textit{validity, explainability, calibration, robustness, governance}, and \textit{cost}. While zero-shot LLMs show modest predictive validity (see Section 4), their self-explanatory faithfulness is lacking (Tables 2-3). This supports the case against direct deployment in regulated finance without safeguards. The results imply that any LLM-based tabular classifier must undergo rigorous validation (e.g. explanation checks, serialization tests), calibration, threshold governance, and human-in-the-loop review before deployment.

\subsubsection{Go/No-Go Deployability Guidance}

\paragraph{Green-lights (pilot only):} 
Use in small-data scenarios where gradient-boosted trees underperform; low-stakes triage; auxiliary signals in a hybrid model (e.g., LLM + LightGBM) with independent tabular baselines and auditing.

\paragraph{Red-flags (no deployment without mitigation):} 
Applications with regulatory or legal consequences (e.g., credit, AML, HR); those requiring faithful feature-level explanations; high prompt formatting sensitivity; failure to pass calibration or explanation sanity tests; absence of monitoring/fallback plans.

\section{Conclusion}
\label{sec:conclousion}

The growing adoption of LLMs for structured classification, especially by non-expert users, raises foundational concerns about their reliability and interpretability. This study systematically evaluates zero-shot LLMs on a suite of financial tabular classification tasks. While results show potential, performance remains modest and explanation fidelity is limited.

Currently, zero-shot LLMs are best viewed as fallback options in small-data settings where fine-tuning is infeasible. Their outputs should not be trusted without rigorous auditing.

Future work should benchmark few-shot and many-shot LLMs using structured and unstructured data. Research into domain-specific fine-tuning and hybrid model integration will be key to making LLMs viable for deployment in high-stakes financial applications.

\bibliography{references}

\section*{Appendix}
\appendix
\section{Features Descriptions}
\label{sec:appendix_features}

Features for all datasets are presented on \Cref{tab:bankruptcy_features,tab:loan_features,tab:license_features}.


\begin{table*}[t]
\centering
\setlength{\tabcolsep}{1mm}
\begin{tabular}{ll}
\toprule
\textbf{Feature description} 
& \textbf{Range} \\
\midrule
((cash + short-term securities + AR - short-term liabilities) / (OPEX - depreciation)) x 365 & [-3039.4, 2104.43] \\
retained earnings / total assets & [-0.72, 0.8] \\
sales / total assets & [0.46, 7.88] \\
(gross profit + extraordinary items + financial expenses) / total assets & [-0.24, 0.84] \\
(gross profit + depreciation) / sales & [-0.18, 0.71] \\
sales (n) / sales (n-1) & [0.53, 2.98] \\
profit on operating activities / total assets & [-0.21, 0.76] \\
gross profit (in 3 years) / total assets & [-0.42, 1.26] \\
(equity - share capital) / total assets & [-0.58, 0.95] \\
(net profit + depreciation) / total liabilities & [-0.32, 8.45] \\
profit on operating activities / financial expenses & [-11.97, 4116.67] \\
logarithm of total assets & [2.9, 6.01] \\
(total liabilities - cash) / sales & [-0.41, 2.22] \\
operating expenses / total liabilities & [-0.27, 27.02] \\
(current assets - inventory) / long-term liabilities & [0.15, 609.35] \\
constant capital / total assets & [-0.19, 0.97] \\
(current assets - inventory - receivables) / short-term liabilities & [0.0, 8.45] \\
net profit / inventory & [-2.68, 46.08] \\
(current assets - inventory) / short-term liabilities & [0.13, 13.6] \\
total costs / total sales & [0.16, 1.23] \\
\bottomrule
\end{tabular}
\caption{Bankruptcy dataset features. AR - account receivables, OPEX - operating expenses. The numerical value intervals are bounded by the 1st and the 99th percentiles for each variable.}
\label{tab:bankruptcy_features}
\end{table*}
\begin{table*}[htb]
\centering
\setlength{\tabcolsep}{1mm}
\begin{tabular}{ll}
\toprule
\textbf{Feature description} 
& \textbf{Range} \\
\midrule
Loan Amount & [1600.0, 35000.0] \\
Term & categorical: \{'36 months', '60 months'\} \\
Interest Rate & [6.03, 25.29] \\
Installment & [55.32, 1204.57] \\
Grade & categorical: \{'A', 'B', 'C', 'D', 'E', 'F', 'G'\} \\
Sub-grade & categorical: \{'A1', 'A2', ... 'G4', 'G5'\} \\
Employment Length & categorical: \{'<1 year', '1 year', '2 years', ..., '10+ years'\} \\
Home Ownership & categorical: \{'MORTGAGE', 'NONE', 'OTHER', 'OWN', 'RENT'\} \\
Annual Income & [19000.0, 250000.0] \\
Verification Status & categorical: \{'Not Verified', 'Source Verified', 'Verified'\} \\
Purpose & categorical: 14 possible values (e.g 'wedding') \\
Debt-to-Income (DTI) Ratio & [1.6, 36.41] \\
Open Credit Accounts & [3.0, 27.0] \\
Public Records & [0.0, 2.0] \\
Revolving Balance & [169.05, 83505.9] \\
Revolving Utilization Rate & [1.2, 98.0] \\
Total Accounts & [6.0, 60.0] \\
Initial Listing Status & categorical: \{'f', 'w'\} \\
Application Type & categorical: \{'DIRECT PAY', 'INDIVIDUAL', 'JOINT'\} \\
Mortgage Accounts & [0.0, 9.0] \\
Public Record Bankruptcies & [0.0, 1.0] \\ 
\bottomrule
\end{tabular}
\caption{Loan Repayment dataset features. The numerical value intervals are bounded by the 1st and the 99th percentiles for each variable. For categorical features, the values space is shown.}
\label{tab:loan_features}
\end{table*}
\begin{table*}[t!]
\centering
\setlength{\tabcolsep}{1mm}
\begin{tabular}{ll}
\toprule
\textbf{Feature description} 
& \textbf{Range} \\
\midrule
Number of unique companies that the employee has worked at & [1.0, 7.0] \\
Number of unique companies that the employee has worked at per working day & [0.0, 0.01] \\
Tenure across all companies (days) & [142.0, 3631.0] \\
Tenure at the current company (days) & [23.0, 3584.0] \\
Longest tenure (days) & [10.0, 2897.69] \\
Average tenure (days) & [7.86, 2744.34] \\
Shortest tenure (days) & [1.0, 2739.07] \\
Gender & \{0.0, 1.0\} \\
Is the employee Hongkonger? & \{0.0, 1.0\} \\
Is the employee Chinese? & \{0.0, 1.0\} \\
Is the employee British? & \{0.0, 1.0\} \\
Number of employees in the company & [2.0, 1669.0] \\
Days of existence of the company & [0.0, 3647.0] \\
Cumulated tenure of all employees in the company & [0.0, 1094284.0] \\
Average tenure in the company & [85.0, 1432.4] \\
\bottomrule
\end{tabular}
\caption{License Expiration dataset features. The numerical value intervals are bounded by the 1st and the 99th percentiles for each variable.}
\label{tab:license_features}
\end{table*}


\end{document}